\documentclass[runningheads]{llncs}
\usepackage{xcolor}
\usepackage{hyperref}
\usepackage{verbatim}
\usepackage{makecell}
\usepackage{graphicx}
\usepackage{colortbl}
\usepackage{amssymb}
\usepackage{amsmath}
\usepackage{pifont}

\usepackage{lipsum}
\usepackage[noend]{algpseudocode}
\usepackage[ruled,vlined]{algorithm2e}
\usepackage{rotating}
\usepackage[export]{adjustbox}
\begin{document}

\title{Self-Supervised Depth Estimation in Laparoscopic Image using 3D Geometric Consistency}

\author{Baoru Huang\inst{1,2} \and
Jian-Qing Zheng\inst{3,4} \and
Anh Nguyen\inst{5}  \and
Chi Xu\inst{1,2} \and
Ioannis Gkouzionis\inst{1,2} \and
Kunal Vyas \inst{6}  \and
David Tuch \inst{6}  \and
Stamatia Giannarou \inst{1,2}  \and
Daniel S. Elson \inst{1,2}}
\institute{The Hamlyn Centre for Robotic Surgery, Imperial College London, London, UK \\
\email{Baoru.Huang18@imperial.ac.uk} \and
Department of Surgery $\&$ Cancer, Imperial College London, London, UK \and
The Kennedy Institute of Rheumatology, University of Oxford, Oxford, UK \and
Big Data Institute, University of Oxford, Oxford, U.K.\and
Department of Computer Science, University of Liverpool, Liverpool, UK \and
Lightpoint Medical Ltd., Chesham, UK\\ }

\maketitle              
\begin{abstract}
Depth estimation is a crucial step for image-guided intervention in robotic surgery and laparoscopic imaging system. Since per-pixel depth ground truth is difficult to acquire for laparoscopic image data, it is rarely possible to apply supervised depth estimation to surgical applications. As an alternative, self-supervised methods have been introduced to train depth estimators using only synchronized stereo image pairs. However, most recent work focused on the left-right consistency in 2D and ignored valuable inherent 3D information on the object in real world coordinates, meaning that the left-right 3D geometric structural consistency is not fully utilized. To overcome this limitation, we present M3Depth, a self-supervised depth estimator to leverage 3D geometric structural information hidden in stereo pairs while keeping monocular inference. The method also removes the influence of border regions unseen in at least one of the stereo images via masking, to enhance the correspondences between left and right images in overlapping areas. Intensive experiments show that our method outperforms previous self-supervised approaches on both a public dataset and a newly acquired dataset by a large margin, indicating a good generalization across different samples and laparoscopes. Code and data are available at \href{https://github.com/br0202/M3Depth}{https://github.com/br0202/M3Depth}. 

\end{abstract}
\section{Introduction}
Perception of 3D surgical scenes is a fundamental problem in computer assisted surgery. Accurate perception, tissue tracking, 3D registration between intra- and pre-operative organ models, target localization and augmented reality \cite{liu2019dense,huang2020tracking} are predicated on having access to correct depth information. Range finding sensors such as multi-camera systems or LiDAR that are often employed in autonomous systems and robotics are not convenient for robot-assisted minimally invasive surgery because of the limited port size and requirement of sterilization. Furthermore, strong `dappled' specular reflections as well as less textured tissues hinder the application of traditional methods \cite{luo2016efficient}. This has led to the exploration of learning-based methods, among which fully convolutional neural networks (CNNs) are particularly successful \cite{allan2021stereo,tran2022light}.   

Since it is challenging to obtain per-pixel ground truth depth for laparoscopic images, there are far fewer datasets in the surgical domain compared with mainstream computer vision applications~\cite{geiger2013vision,huang2022h}. It is also not a trivial task to transfer approaches that are based on supervised learning to laparoscopic applications due to the domain gap. To overcome these limitations, view-synthesis methods are proposed to provide self-supervised learning for depth estimation \cite{huang2021self,liu2019dense}, with no supervision via per-pixel depth data. Strong depth prediction baselines have been established in \cite{godard2017unsupervised,godard2019digging,johnston2020self}. However, all of these methodologies employed left-right consistency and smoothness constraints in 2D, \textit{e.g.} \cite{godard2017unsupervised,guizilini20203d,huang2022simultaneous}, 
and ignored the important 3D geometric structural consistency from the stereo images.

Recently, a self-supervised semantically-guided depth estimation method was proposed to deal with moving objects \cite{klingner2020self}, which made use of mutually beneficial cross-domain training of semantic segmentation. Jung \textit{et al.} \cite{jung2021fine} extended this work by incorporating semantics-guided local geometry into intermediate depth representations for geometric representation enhancement. However, semantic labels are not common in laparoscopic applications except for surgical tool masks, impeding the extension of this work. Mahjourian \textit{et al.} \cite{mahjourian2018unsupervised} presented an approach for unsupervised learning of depth by enforcing consistency of ego-motion across consecutive frames to infer 3D geometry of the whole scene. However, in laparoscopic applications, the interaction between the surgical tools and tissue creates a dynamic scene, leading to failure of local photometric and geometric consistency across consecutive frames in both 2D and 3D. Nevertheless, the 3D geometry inferred from left and right synchronized images can be assumed identical, allowing adoption of 3D- as well as 2D-loss.   

In this paper, we propose a new framework for self-supervised laparoscopic image depth estimation called M3Depth, leveraging not only the left-right consistency in 2D but also the inherent geometric structural consistency of real-world objects in 3D (see section \ref{3dloss} for the 3D geometric consistency loss), while enhancing the mutual information between stereo pairs. A U-Net architecture \cite{ronneberger2015u} was employed as the backbone and the network was fed with only left image as inputs but was trained  with the punitive loss formed by stereo image pairs. To cope with the unseen areas at the image edges that were not visible in both cameras, blind masking was applied to suppress and eliminate outliers and give more emphasis to feature correspondences that lay on the shared vision field. 
Extensive experiments on both a public dataset and a new experimental dataset demonstrated the effectiveness of this approach and a detailed ablation study indicated the respective positive influence of each proposed novel module on the overall performance.

\section{Methodology}

\subsection{Network Architecture}

\begin{figure}[]
\includegraphics[width=\textwidth, height=0.45\textwidth]{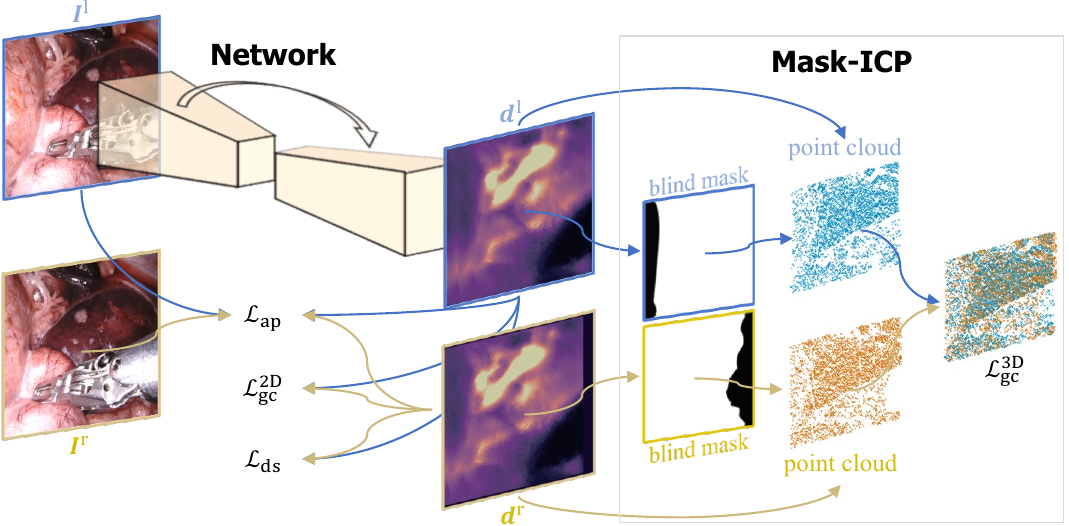}
\caption{Overview of the proposed self-supervised depth estimation network. ResNet18 was adopted as the backbone and received a left image from a stereo image pair as the input. Left and right disparity maps were produced simultaneously and formed 2D losses with the original stereo pair. 3D point clouds were generated by applying the intrinsic parameters of the camera and iterative closest point loss was calculated between them. Blind masks were applied to the 2D disparity maps to remove outliers from areas not visible in both cameras. }

\label{overview}
\end{figure}

\subsubsection{Network Architecture}
The backbone of the M3Depth followed the general U-Net \cite{ronneberger2015u} architecture, \textit{i.e.} an encoder-decoder network, in which an encoder was employed to extract image representations while a decoder with convolutional layer and upsampling manipulation was designed to recover disparity maps at the original scale. Skip connections were applied to obtain both deep abstract features and local information. To keep a lightweight network, a ResNet18 \cite{he2016deep} was employed as the encoder with only 11 million parameters. To improve the regression ability of the network from intermediate high-dimensional features maps to disparity maps, one more ReLU \cite{nair2010rectified} activation function and a convolutional layer with decreased last latent feature map dimension were added before the final sigmoid disparity prediction. Similar to Monodepth1 \cite{godard2017unsupervised}, in M3Depth, the left image ${\textbf{\textit{I}}^{\rm l}}$ of a stereo image pair (${\textbf{\textit{I}}^{\rm l}},{\textbf{\textit{I}}^{\rm r}}\in\mathbb{R}_{+}^{h\times w\times 3}$) was always the input and the framework generated two distinct left and right disparity maps ${\textbf{\textit{d}}^{\rm l}}$, ${\textbf{\textit{d}}^{\rm r}} \in\mathbb{R}_{+}^{h\times w}$ simultaneously, \textit{i.e.}  ${\mathcal{Z}: \textbf{\textit{I}}^{\rm l}\mapsto({\textbf{\textit{d}}^{\rm l}},{\textbf{\textit{d}}^{\rm r}})}$. Given the camera focal length $\textbf{\textit{f}}$ and the baseline distance $\textbf{\textit{b}}$ between the cameras, left and right depth maps $\textbf{\textit{D}}^{\rm l},\textbf{\textit{D}}^{\rm r}\in\mathbb{R}_{+}^{h\times w}$ could then be trivially recovered from the predicted disparity, ($\textbf{\textit{D}}^{\rm l},\textbf{\textit{D}}^{\rm r}$)=$\textbf{\textit{bf}}$/(${\textbf{\textit{d}}^{\rm l}}$, ${\textbf{\textit{d}}^{\rm r}}$). $h$ and $w$ denote image height and width. Full details of the architecture are presented in the supplementary material.

\subsubsection{Image Reconstruction Loss in 2D}
With the predicted disparity maps and the original stereo image pair, left and right images could then be reconstructed by warping the counter-part RGB image with the disparity map mimicking optical flow \cite{godard2017unsupervised} \cite{lipson2021raft}. Similar to Monodepth1 \cite{godard2017unsupervised}, an appearance matching loss \(\mathcal{L}_{\rm ap}\), disparity smoothness loss \(\mathcal{L}_{\rm ds}\) and left-right disparity consistency loss \(\mathcal{L}_{\rm lr}^{2D}\) were used to encourage coherence between the original input and reconstructed images (\(\textbf{\textit{I}}^{\rm l*}, \textbf{\textit{I}}^{\rm r*}\)) as well as consistency between left and right disparities while forcing disparity maps to be locally smooth.

\begin{equation} \label{eq:appearance_matching_loss}
\mathcal{L}_{ap}^{\rm r} = \frac{1}{N}\sum_{i, j} \frac{\gamma}{2} (1-{\rm SSIM}(I_{ij}^{\rm r}, I_{ij}^{r*})) + (1-\gamma) {\|I_{ij}^{\rm r}-I_{ij}^{r*}\|}_1
\end{equation}

\begin{equation}
\mathcal{L}_{\rm lr}^{\rm 2D(r)} = \frac{1}{N}\sum_{ij}| \textbf{d}_{ij}^{\rm r} +  \textbf{d}_{ij+\textbf{d}_{ij}^{\rm r}}^{\rm l}|
\end{equation}

\begin{equation}
\mathcal{L}_{\rm ds}^{\rm r} = \frac{1}{N}\sum_{ij}|\partial_x (\textbf{d}_{ij}^{\rm r})|e^{-|\partial_x \textbf{\textit{I}}_{ij}^{\rm r}|} + |\partial_y (\textbf{d}_{ij}^{\rm r})| e^{-|\partial_y \textbf{\textit{I}}_{ij}^{\rm r}|}
\end{equation}
where \(N\) is the number of pixels and \(\gamma\) was set to 0.85. Note that 2D losses were applied on both left and right images but only equations for the right image are presented here.

\subsection{Learning 3D Geometric Consistency}
\label{3dloss}
Instead of using the inferred left and right disparities only to establish a mapping between stereo coordinates and generate reconstructed original RGB input images, a loss function was also constructed that registered and compared left and right point clouds directly to enforce the 3D geometric consistency of the whole scene. The disparity maps of the left and right images were first converted to depth maps and then backprojected to 3D coordinates to obtain left and right surgical scene point clouds ($\textbf{\textit{P}}^{\rm l},\textbf{\textit{P}}^{\rm r}\in\mathbb{R}^{hw\times3}$) by multiplying the depth maps with the intrinsic parameter matrix (\(\textbf{\textit{K}}\)). The 3D consistency loss employed Iterative Closest Point (ICP) \cite{mahjourian2018unsupervised,rusinkiewicz2001efficient}, a classic rigid registration method that derives a transformation matrix between two point clouds by iteratively minimizing point-to-point distances between correspondences. 

From an initial alignment, ICP alternately computed corresponding points between two input point clouds using a closest point heuristic and then recomputed a more accurate transformation based on the given correspondences. The final residual registration error after ICP minimization was output as one of the returned values. More specifically, to explicitly explore global 3D loss, the ICP loss at the original input image scale was calculated with only 1000 randomly selected points to reduce the computational workload. 

\subsection{Blind Masking}
Some parts of the left scene were not visible in the right view and vice versa, leading to non-overlapping generated point clouds. These areas are mainly located at the left edge of the left image and right edge of the right image after rectification. Depth and image pixels in those area had no useful information for learning, either in 2D and 3D. Our experiments indicated that retaining the contribution to the loss functions for such pixels and voxels degraded the overall performance. Many previous approaches solved this problem by padding these areas with zeros \cite{godard2017unsupervised} or values from the border \cite{godard2019digging}, but this can lead to edge artifacts in depth images~\cite{mahjourian2018unsupervised}.

To tackle this problem, we present a blind masking module \(\mathcal{M}^{\rm l,r}\) that suppressed and eliminated these outliers and gave more emphasis to correspondences between the left and right views. First, a meshgrid was built with the original left image pixel coordinates in both horizontal \(\mathcal{X}_{\rm grid}\) and vertical \(\mathcal{Y}_{\rm grid}\) directions. Then the \(\mathcal{X}_{\rm grid}\) was shifted along the horizontal direction using the right disparity map ${\textbf{\textit{d}}^{\rm r}}$ to a get a new grid  \(\mathcal{X'}_{\rm grid}\), which was then stacked with \(\mathcal{Y}_{\rm grid}\) to form a new meshgrid. Finally, grid sampling was employed on the new meshgrid with the help of the original left image coordinates, from which the pixels that were not covered by the right view for the current synchronized image pair were obtained. By applying the blind masking on the depth maps for the stereo 3D point cloud generation, a 3D alignment loss was obtained as follows. 

\begin{equation}
\textbf{\textit{M}}_{ij}^{\rm l,r} =  \left\{
\begin{array}{cc}
1, & (\textbf{d}_{ij}^{\rm l,r}+\mathcal{X}_{ij})\in\{\mathcal{X}\}  \\
0, & (\textbf{d}_{ij}^{\rm l,r}+\mathcal{X}_{ij})\notin\{\mathcal{X}\}
\end{array}
\right.
\end{equation}
\begin{equation}
\textbf{\textit{P}}^{\rm l,r}={\rm backproj}(\textbf{d}^{\rm l,r},\textbf{\textit{K}},\textbf{\textit{M}}^{\rm l,r})
\end{equation}
\begin{equation}
\mathcal{L}_{\rm gc}^{\rm 3D} = 
{\rm ICP}(\textbf{\textit{P}}^{\rm l}, \textbf{\textit{P}}^{\rm r})
\end{equation}

\subsection{Training Loss}
Pixel-wise, gradient-based 2D losses and point cloud-based 3D losses were applied to force the reconstructed image to be identical to the original input while encouraging the left-right consistency in both 2D and 3D to derive accurate disparity maps for depth inference. Finally, an optimization loss used a combination of these, written as:

\begin{equation}
\begin{split}
\mathcal{L}_{\rm total} & = (\mathcal{L}_{\rm 2D}^{\rm r} + \mathcal{L}_{\rm 2D}^{\rm l}) +  \mathcal{L}_{\rm lr}^{\rm 3D} \\
& = \alpha_{\rm ap}(\mathcal{L}_{\rm ap}^{\rm r} + \mathcal{L}_{\rm ap}^{\rm l}) + \alpha_{\rm ds}( \mathcal{L}_{\rm ds}^{\rm r} + \mathcal{L}_{\rm ds}^{\rm l}) + \alpha_{\rm lr}^{2D}(\mathcal{L}_{\rm lr}^{\rm 2D(r)} + \mathcal{L}_{\rm lr}^{\rm 2D(l)}) + \beta \mathcal{L}_{\rm gc}^{\rm 3D}
\end{split}
\end{equation}
where \(\alpha_{\rm *}\) and \(\beta\) balanced the loss magnitude of the 2D and 3D parts to stabilize the training. More specifically, \(\alpha_{\rm ap}\), \(\alpha_{\rm ds}\), \(\alpha_{\rm lr}^{\rm 2D}\) and \(\beta\) were experimentally set to 1.0, 0.5, 1.0 and 0.001.

\section{Experiments}
\subsection{Dataset}

M3Depth was evaluated on two datasets. The first was the \textit{SCARED} dataset \cite{allan2021stereo} released at the MICCAI Endovis challenge 2019. As only the ground truth depth map of key frames in each dataset was provided (from structured light), the other depth maps were created by reprojection and interpolation of the key frame depth maps using the kinematic information from the \textit{da Vinci} robot, causing a misalignment between the ground truth and the RGB data. Hence, only key-frame ground truth depth maps were used in the test dataset while the remainder of the RGB data formed the training set but with the similar adjacent frames removed. 

To overcome the \textit{SCARED} dataset misalignment and improve the validation, an additional laparoscopic image dataset (namely \textit{LATTE}) was experimentally collected, including RGB laparoscopic images and corresponding ground truth depth maps calculated from a custom-built structured lighting pattern projection. More specifically, the gray-code detection and decoder algorithm \cite{xu2007robust} were used with both original and inverse patterns. To remove the uncertainty brought by occlusions and uneven illumination conditions, we used a more advanced 3-phase detection module, in which sine waves were shifted by $\pi/3$ and $2\pi/3$ and the modulation depth $\mathcal{T}$ was calculated for every pixel. Pixels with modulation depth under $\mathcal{T}$ were defined as uncertain pixels, and the equation for calculating the modulation is written as Eq.~\ref{eq:modulation}. This provided 739 extra image pairs for training and 100 pairs for validation and testing.

\begin{equation}
\mathcal{T} = \frac{2\sqrt{2 }}{3}\times\sqrt{(\mathcal{I}_{1} - \mathcal{I}_{2})^2 + (\mathcal{I}_{2} - \mathcal{I}_{3})^2 + (\mathcal{I}_{1} - \mathcal{I}_{3})^2}
\label{eq:modulation}
\end{equation}
where $\mathcal{I}_{1}, \mathcal{I}_{2}, \mathcal{I}_{3}$ denotes original modulation and modulations after shifts.

\begin{figure*}[!t]
  \centering
  \includegraphics[width=\textwidth]{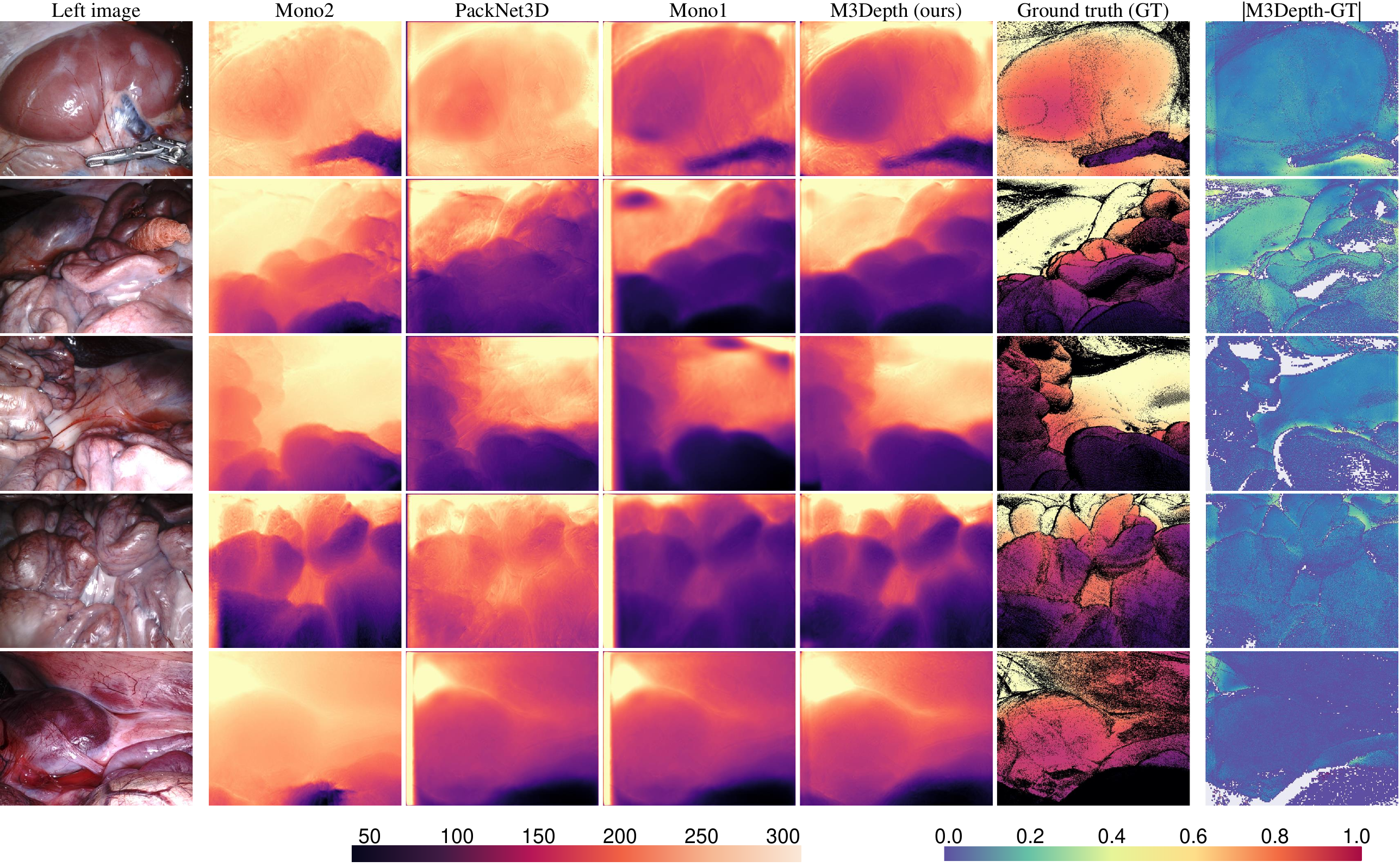}
  
  \caption{Qualitative results on the \textit{SCARED} dataset with error map of M3Depth. The depth predictions are all for the left input image. M3Depth generated depth maps with high contrast between the foreground and background and performed better at distinguishing different parts of the scene, reflecting the superior quantitative results in Table \ref{quantitive results SCARED}.}
  \label{Fig:visualization}
\end{figure*}

\begin{table*}[h]
\caption{Quantitative results on the \textit{SCARED} dataset. 
Metrics labeled with blue headings mean \textit{lower is better} while those labeled with red mean \textit{higher is better}.
}
\resizebox{\textwidth}{!}{
\begin{tabular}{|c|c|c|c|c|c|c|c|c|}
\hline

\hline
Method   & \cellcolor[RGB]{193,218,243}Abs Rel & \cellcolor[RGB]{193,218,243}Sq Rel & \cellcolor[RGB]{193,218,243}RMSE  & \cellcolor[RGB]{193,218,243}RMSE log & \cellcolor[RGB]{249,196,196}${\delta<1.25 }$ &\cellcolor[RGB]{249,196,196} $\delta<1.25^2$  &\cellcolor[RGB]{249,196,196}${\delta<1.25^3}$ \\ 
\hline

Mono2 \cite{godard2019digging}  &1.100 &74.408 &56.548 &0.717 &0.102 &0.284 &0.476\\

PackNet3D \cite{guizilini20203d} &   0.733  &  37.690  &  32.579  &   0.649  &   0.288  &   0.538  &   0.722  \\

Mono1 \cite{godard2017unsupervised} &0.257  &20.649  &33.796  &0.404  &0.696  &0.837  &0.877 \\

\rowcolor[RGB]{230,230,230}
\textbf{M3Depth}  
&   \textbf{0.116}  &   \textbf{1.822}  &   \textbf{9.274}  &   \textbf{0.139}  &   \textbf{0.865}  &   \textbf{0.983}  &   \textbf{0.997} \\

\hline

\hline
\end{tabular}}
\label{quantitive results SCARED}
\end{table*}

\begin{table*}[h]
\caption{Quantitative results on \textit{LATTE}  dataset. 
Metrics labeled with blue headings mean \textit{lower is better} while labeled by red mean \textit{higher is better}.
}
\resizebox{\textwidth}{!}{
\begin{tabular}{|c|c|c|c|c|c|c|c|c|}
\hline

\hline
Method   & \cellcolor[RGB]{193,218,243}Abs Rel & \cellcolor[RGB]{193,218,243}Sq Rel & \cellcolor[RGB]{193,218,243}RMSE  & \cellcolor[RGB]{193,218,243}RMSE log & \cellcolor[RGB]{249,196,196}${\delta<1.25 }$ &\cellcolor[RGB]{249,196,196} $\delta<1.25^2$  &\cellcolor[RGB]{249,196,196}${\delta<1.25^3}$ \\ 
\hline

Mono2 \cite{godard2019digging}     &   1.601  & 306.823  &  87.694  &   0.913  &   0.169  &   0.384  &   0.620   \\

PackNet3D \cite{guizilini20203d} &   0.960  & 357.023  & 259.627  &   0.669  &   0.135  &   0.383  &   0.624  \\

Mono1 \cite{godard2017unsupervised} 
&   0.389  &  57.513  &  99.020  &   0.424  &   0.268  &   0.709  &   0.934   \\

\rowcolor[RGB]{230,230,230}
\textbf{M3Depth}    &   \textbf{0.236}  &  \textbf{21.839}  &  \textbf{57.739}  &   \textbf{0.245}  &   \textbf{0.665}  &   \textbf{0.893}  &   \textbf{0.969} \\

\hline

\hline
\end{tabular}}
\label{quantitive results on LATTE}
\end{table*}

\subsection{Evaluation Metrics, Baseline, and Implementation Details}
\subsubsection{Evaluation Metrics} To evaluate depth errors, seven criteria were adopted that are commonly used for monocular depth estimation tasks \cite{godard2017unsupervised}\cite{godard2019digging}: mean absolute error (Abs Rel), squared error (Sq Rel), root mean squared error (RMSE), root mean squared logarithmic error (RMSE log), and the ratio between ground truth and prediction values, for which the threshold was denoted as ${\delta}$.

\subsubsection{Baseline}
The M3Depth model was compared with several recent deep learning methods including Monodepth \cite{godard2017unsupervised}, Monodepth2 \cite{godard2019digging}, and PackNet \cite{guizilini20203d}, and both quantitative and qualitative results were generated and reported for comparison. To further study the importance of each M3Depth component, the various components of M3Depth were removed in turn.

\subsubsection{Implementation Details}
M3Depth was implemented in PyTorch \cite{paszke2017automatic}, with an input/output resolution of \(256\times320\) and a batch size of 18. The learning rate was initially set to \(10^{-5}\) for the first 30 epochs and was then halved until the end. The model was trained for 50 epochs using the Adam optimizer which took about 65 hours on two NVIDIA 2080 Ti GPUs.

\section{Results and Discussion}

\begin{table*}[h]
\caption{Ablation study results on the \textit{SCARED} dataset. 
Metrics labeled with blue headings mean \textit{lower is better} while those labeled with red mean \textit{higher is better}.
}
\resizebox{\textwidth}{!}{
\begin{tabular}{|c|c|c|c|c|c|c|c|c|c|c|}
\hline

\hline

Method &3GC &Blind masking  & \cellcolor[RGB]{193,218,243}Abs Rel & \cellcolor[RGB]{193,218,243}Sq Rel & \cellcolor[RGB]{193,218,243}RMSE  & \cellcolor[RGB]{193,218,243}RMSE log & \cellcolor[RGB]{249,196,196}${\delta<1.25 }$ &\cellcolor[RGB]{249,196,196} $\delta<1.25^2$  &\cellcolor[RGB]{249,196,196}${\delta<1.25^3}$ \\ 
\hline

Mono1 \cite{godard2017unsupervised}  &\ding{55} &\ding{55}     &0.257  &20.649  &33.796  &0.404  &0.696  &0.837  &0.877 \\

M3Depth w/ mask &\checkmark &\ding{55}   &   0.150  &   3.069  &  13.671  &   0.249  &   0.754  &   0.910  &   0.956  \\
 
\rowcolor[RGB]{230,230,230}
\textbf{M3Depth}  &\checkmark &\checkmark 
&   \textbf{0.116}  &   \textbf{1.822}  &   \textbf{9.274}  &   \textbf{0.139}  &   \textbf{0.865}  &   \textbf{0.983}  &   \textbf{0.997} \\

\hline

\hline
\end{tabular}}
\label{ablation results on SCARED dataset}
\end{table*}

The M3Depth and other state-of-the-art results on the \textit{SCARED} and \textit{LATTE} dataset are shown in Table \ref{quantitive results SCARED} and Table \ref{quantitive results on LATTE} using the seven criteria evaluation metrics. M3Depth outperformed all other methods by a large margin on all seven criteria, which shows that taking the 3D structure of the world into consideration benefited the overall performance of the depth estimation task. In particular, the M3Depth model had 0.141, 18.827, 24.522, and 0.265 units error lower than Monodepth1\cite{godard2017unsupervised} in Abs Rel, Sq Rel, RMSE and RMSE log, and 0.169, 0.146, and 0.12 units higher than Monodepth1 \cite{godard2017unsupervised} in three different threshold criteria. Furthermore, the average inference time of M3Depth was 105 frames per second, satisfying the real-time depth map generation requirements.

Detailed ablation study results on the \textit{SCARED} dataset are shown in Table~\ref{ablation results on SCARED dataset} and the impact of the proposed modules, 3D geometric consistency (3GC) and blind masking were evaluated. The evaluation measures steadily improved when the various components were added. More specifically, the addition of the blind masking further boosted the 3GC term, which shows the importance and necessity of removing invalid information from areas that are not visible to both cameras. We note that more quantitative results can be found in our supplementary material.

Qualitative results comparing our depth estimation results against prior work using the \textit{SCARED} dataset are presented in Fig \ref{Fig:visualization}.
As the sample images shows, the application of temporal consistency encouraged by the 3D geometric consistency loss can reduce the errors caused by subsurface features, and better recover the real 3D surface shape of the tissue. Furthermore, depth outputs from M3Depth show better results along the boundaries of objects, indicating the effectiveness of the proposed 3GC and blind masking modules.

\section{Conclusion}
A novel framework for self-supervised monocular laparoscopic images depth estimation was presented. By combining the 2D image-based losses and 3D geometry-based losses from an inferred 3D point cloud of the whole scene, the global consistency and small local neighborhoods were both explicitly taken into consideration. Incorporation of blind masking avoided penalizing areas where no useful information exists. The modules proposed can easily be plugged into any similar depth estimation network, monocular and stereo, while the use of a lightweight ResNet18 backbone enabled real time depth map generation in laparoscopic applications. Extensive experiments on both public and newly acquired datasets demonstrated good generalization across different laparoscopes, illumination condition and samples, indicating the capability to large scale data acquisition where precise ground truth depth cannot be easily collected.

\bibliographystyle{splncs04}
\bibliography{paper}

\title{Supplementary material}

\author{}
\institute{}
\maketitle              

\section{Network Architecture}

\begin{figure}[]
\includegraphics[width=\textwidth]{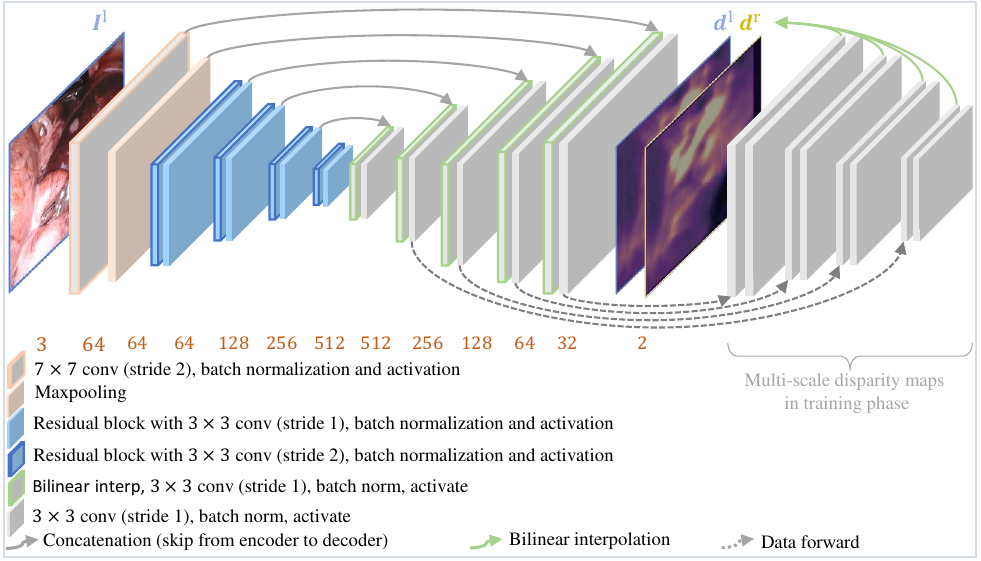}
\caption{
The detailed architecture of the M3Depth framework. A U-Net \cite{ronneberger2015u} structure with ResNet18 \cite{he2016deep} lightweight network was used as the backbone. The M3Depth took a left image as an input and simultaneously generated left and right disparity maps. 2D and 3D losses were formed by combining the disparities of output and original stereo pairs, which were used as constraints for the whole network training. Loss details are shown in the main paper.}

\label{network}
\end{figure}


\clearpage
\section{Data Acquisition of \textit{LATTE}}

\begin{figure}[]
\includegraphics[width=\textwidth]{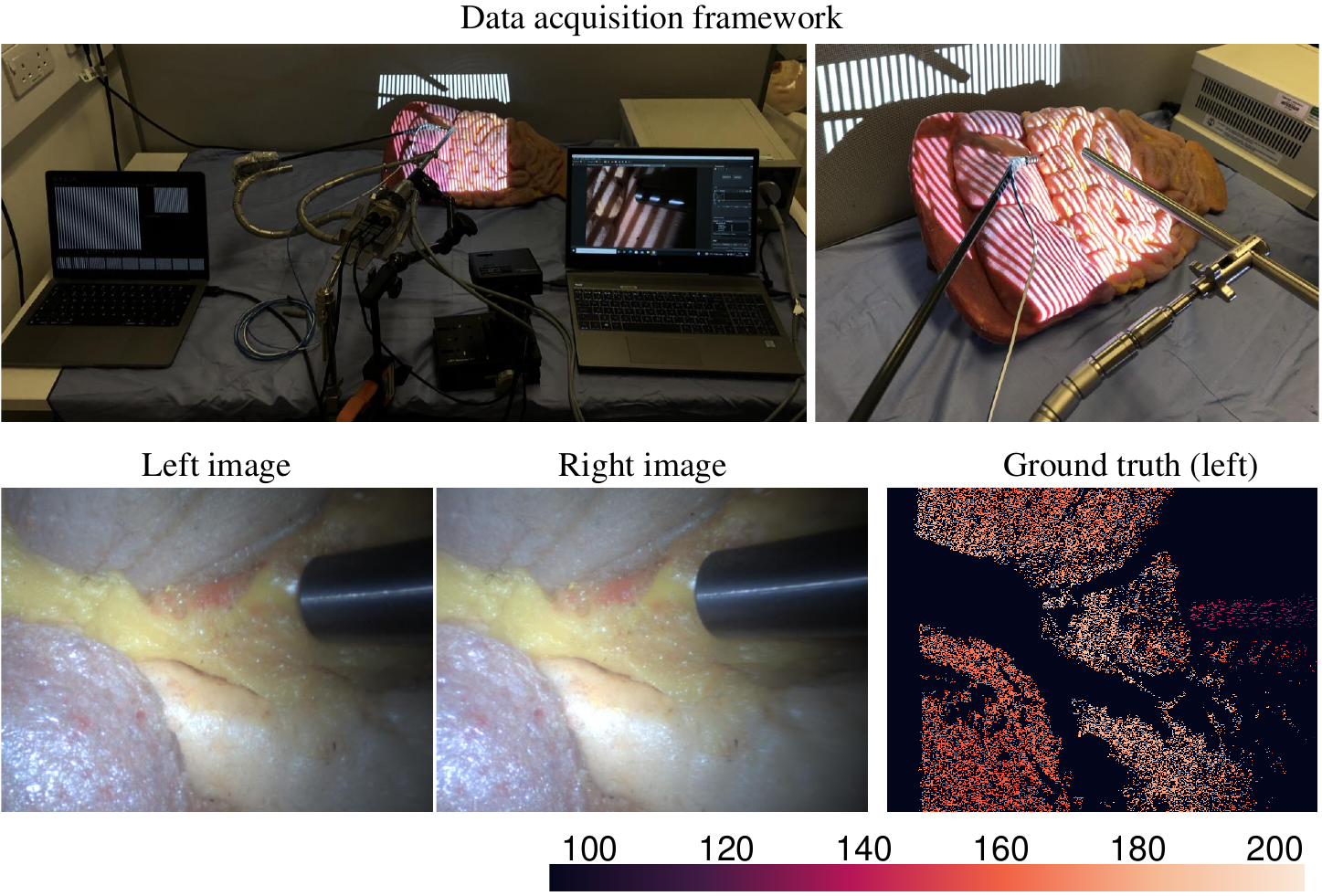}
\caption{
The data acquisition setting and dataset example of \textit{LATTE}.}

\label{network}
\end{figure}

\clearpage

\section{Further Ablation Study}
\begin{table*}[h]
\caption{Further ablation study results on the \textit{SCARED} dataset for the 3D geometric consistency (3GC), blind masking (BM), and fine disp (FD) modules. Please note that fine disp indicates the modification of the decoder used in Monodepth 1 \cite{godard2017unsupervised}, including the addition of one more ReLU \cite{nair2010rectified} activation functions and a convolutional layer with decreased last latent feature map dimension. 
Metrics that are labeled with blue headings mean \textit{lower is better} while those labeled with red mean \textit{higher is better}.
}
\resizebox{\textwidth}{!}{
\begin{tabular}{|c|c|c|c|c|c|c|c|c|c|c|c|}
\hline

\hline

Method &3GC &BM &FD  & \cellcolor[RGB]{193,218,243}Abs Rel & \cellcolor[RGB]{193,218,243}Sq Rel & \cellcolor[RGB]{193,218,243}RMSE  & \cellcolor[RGB]{193,218,243}RMSE log & \cellcolor[RGB]{249,196,196}${\delta<1.25 }$ &\cellcolor[RGB]{249,196,196} $\delta<1.25^2$  &\cellcolor[RGB]{249,196,196}${\delta<1.25^3}$ \\ 
\hline

Mono1 \cite{godard2017unsupervised} &\ding{55} &\ding{55}  &\ding{55}   &0.257  &20.649  &33.796  &0.404  &0.696  &0.837  &0.877 \\

Mono1 w/ 3GC &\checkmark &\ding{55}  &\ding{55}   &   0.125  &   1.880  &   9.625  &   0.179  &   0.822  &   0.955  &   0.980 \\

Mono1 w/ 3GC, BM &\checkmark &\checkmark  &\ding{55}   &0.121  &2.078  &9.723 &0.148  &0.842  &0.977  &0.994  \\

Mono1 w/ FD &\ding{55} &\ding{55} &\checkmark &   0.120  &   2.154  &  10.071  &   0.153  &   0.850  &   0.973  &   0.994  \\

Mono1 w/ 3GC, FD &\checkmark &\ding{55} &\checkmark  &   0.150  &   3.069  &  13.671  &   0.249  &   0.754  &   0.910  &   0.956  \\
 
\rowcolor[RGB]{230,230,230}
\textbf{M3Depth}  &\checkmark &\checkmark &\checkmark
&   \textbf{0.116}  &   \textbf{1.822}  &   \textbf{9.274}  &   \textbf{0.139}  &   \textbf{0.865}  &   \textbf{0.983}  &   \textbf{0.997} \\

\hline

\hline
\end{tabular}}
\label{ablation results on SCARED dataset}
\end{table*}

\begin{table*}[h]
\caption{Quantitative results on the \textit{SCARED} dataset using the split of MICCAI challenge and test on the keyframes of the test dataset. The geometrically oriented mono version of Monodepth2 \cite{godard2019digging} is marked with `Geo'.
Metrics labeled with blue headings mean \textit{lower is better} while those labeled with red mean \textit{higher is better}.
}
\resizebox{\textwidth}{!}{
\begin{tabular}{|c|c|c|c|c|c|c|c|c|}
\hline

\hline
Method   & \cellcolor[RGB]{193,218,243}Abs Rel & \cellcolor[RGB]{193,218,243}Sq Rel & \cellcolor[RGB]{193,218,243}RMSE  & \cellcolor[RGB]{193,218,243}RMSE log & \cellcolor[RGB]{249,196,196}${\delta<1.25 }$ &\cellcolor[RGB]{249,196,196} $\delta<1.25^2$  &\cellcolor[RGB]{249,196,196}${\delta<1.25^3}$ \\ 
\hline

Mono2-Geo \cite{godard2019digging} &0.335  &10.999  &25.441  &0.434  &0.248  &0.662  &0.831 \\
Mono1 \cite{godard2017unsupervised} &0.328  & 16.973  & 31.031  & 0.401  &  0.388  &  0.700  & \textbf{0.901}  \\

Bian \textit{et al.} \cite{bian2019unsupervised} &0.302  &9.207  &22.427  &0.371  &0.429  &0.701  &0.848 \\ 

\rowcolor[RGB]{230,230,230}
\textbf{M3Depth}  
&\textbf{0.274} &  \textbf{8.216}  &  \textbf{20.678}  & \textbf{0.350}  & \textbf{0.470}  &  \textbf{0.723}  &  0.888 \\

\hline

\hline
\end{tabular}}
\label{quantitive results SCARED}
\end{table*}

\end{document}